\documentclass{article}
\usepackage{ijcai26}

\usepackage{times}
\usepackage{soul}
\usepackage{url}
\usepackage[hidelinks]{hyperref}
\usepackage[utf8]{inputenc}
\usepackage[small]{caption}
\usepackage{graphicx}
\usepackage{amsmath}
\usepackage{amsthm}
\usepackage{booktabs}
\usepackage{threeparttable}
\usepackage{algorithm}
\usepackage{algpseudocode}
\usepackage[switch]{lineno}

\usepackage{amsmath,amssymb,amsfonts}
\usepackage{graphicx}
\usepackage{booktabs}
\usepackage{textcomp}
\usepackage[most]{tcolorbox}
\usepackage[table,xcdraw]{xcolor}
\usepackage{caption}

\definecolor{lightblueframe}{HTML}{99CCFF}
\definecolor{lightorange}{HTML}{FF9966}

\title{LLM as Clinical Graph Structure Refiner: Enhancing Representation Learning in EEG Seizure Diagnosis}

\author{
Lincan Li$^1$
\and
Zheng Chen$^2$\and
Yushun Dong$^{1}$\thanks{Yushun Dong is the corresponding author.}\\
\affiliations
$^1$Department of Computer Science, Florida State University,
$^2$SANKEN, The University of Osaka\\
\emails
\{ll24bb,yushun.dong\}@fsu.edu,
chenz@sanken.osaka-u.ac.jp
}

\begin{document}

\maketitle

\begin{abstract}
Electroencephalogram (EEG) signals are vital for automated seizure detection, but their inherent noise makes robust representation learning challenging. Existing graph construction methods, whether correlation-based or learning-based, often generate redundant or irrelevant edges due to the noisy nature of EEG data. This significantly impairs the quality of graph representation and limits downstream task performance. Motivated by the remarkable reasoning and contextual understanding capabilities of large language models (LLMs), we explore the idea of using LLMs as graph edge refiners. Specifically, we propose a two-stage framework: we first verify that LLM-based edge refinement can effectively identify and remove redundant connections, leading to significant improvements in seizure detection accuracy and more meaningful graph structures. Building on this insight, we further develop a robust solution where the initial graph is constructed using a Transformer-based edge predictor and multilayer perceptron, assigning probability scores to potential edges and applying a threshold to determine their existence. The LLM then acts as an edge set refiner, making informed decisions based on both textual and statistical features of node pairs to validate the remaining connections. Extensive experiments on TUSZ dataset demonstrate that our LLM-refined graph learning framework not only enhances task performance but also yields cleaner and more interpretable graph representations. 
\end{abstract}

\section{Introduction}
Accurate epileptic seizure detection is a critical task in both clinical and research fields. Epilepsy affects millions of people worldwide, and accurate identification of seizure events is essential for timely diagnosis, effective treatment, and improved patient quality of life~\cite{begley2022global}. Electroencephalogram (EEG) signals are the primary data source for monitoring brain activities, playing a central role in epilepsy diagnosis and treatment~\cite{benbadis2020role}. EEG records electrical activity across multiple brain regions using an array of electrodes, providing valuable information about both normal and abnormal brain states. However, analyzing EEG data remains a significant challenge, as it is usually contaminated by noise, artifacts, and inter-patient variability~\cite{saba2021unsupervised,aldahr2022addressing}. Furthermore, seizure events vary greatly in duration, frequency, and spatial manifestation across individuals. Traditional EEG analysis methods, such as manual inspection or basic signal processing techniques, are time-consuming and prone to human error~\cite{rani2024attcnnnet}. While recent deep learning approaches have improved performance, they usually struggle to effectively model the rich semantic context and higher-order relationships between EEG channels. These limitations make it difficult to distinguish meaningful patterns from noise or redundancy, highlighting the need for more advanced techniques with stronger reasoning and contextual understanding capabilities, which large language models possess.

Specifically, existing graph learning-based EEG seizure detection methods mainly fall into two categories: those that construct graphs using pairwise correlation or similarity between EEG channels, and those that adopt dynamic or data-driven approaches to learn graph structures from data~\cite{ho2023self,tang2022selfsupervised,pmlr-v209-tang23a}. In the former, the graph is typically predefined based on statistical measures such as normalized cross-correlation or mutual information between channel pairs, and then kept fixed throughout model training~\cite{ho2023self,tang2022selfsupervised}. In the latter, more recent works attempt to learn or adapt graph connectivity directly from the data using neural network-based mechanisms, such as attention modules or state space models~\cite{pmlr-v209-tang23a,dong2023towards}. However, both categories suffer from the same fundamental limitation: the noisy and variable nature of EEG signals makes it difficult to construct accurate and robust graphs. Whether based on correlation, similarity, or data-driven learning, the resulting graphs often contain redundant, spurious, or even erroneous edges, which can mislead graph neural networks (GNNs) and degrade both representation learning and seizure detection performance. Although some methods have attempted to alleviate this issue, most still lack explicit and principled mechanisms to refine graph connectivity and remove task-irrelevant connections. In contrast, LLM-based graph structure learning offers powerful reasoning and contextual understanding, enabling more intelligent and interpretable edge refinement that addresses the core weaknesses of previous approaches.

Therefore, improving the quality of graph structures becomes essential for advancing EEG seizure detection. An effective graph should capture true neural interactions while suppressing the redundant or noisy connections that mislead graph representation learning~\cite{yu2026healthmamba}. Nevertheless, most existing approaches lack the structural learning and semantic understanding abilities needed to refine these connections. Large language models (LLMs) have demonstrated remarkable capabilities in reasoning, contextual understanding, and integrating diverse information sources~\cite{cui2024neuro}. These strengths make LLMs well-suited for graph edge refinement, as they can more effectively distinguish between meaningful and irrelevant connections based on both statistical and semantic evidence. By leveraging LLMs for graph structure learning, it becomes possible to construct graphs that are not only more accurate and robust, but also more interpretable and aligned with underlying physiological mechanisms. This motivates our work to explore LLM-based edge refinement as a promising direction for addressing the key limitations of existing methods in EEG seizure detection. Our main contributions are summarized in the following:

\begin{itemize}
\item We propose a novel framework that leverages large language models (LLMs) as tools for clinical graph structure refinement, overcoming the limitations of existing methods in clinical graph representation learning and contextual understanding.


\item We establish a comprehensive benchmark for LLM-based clinical graph structure refinement, systematically evaluating a range of state-of-the-art general-purpose LLMs as structural judges. This benchmark reveals the varying reasoning capabilities, consistency, and robustness of different LLMs in clinical graph reasoning.

\item Comprehensive experimental evaluations are conducted on TUSZ benchmark, demonstrating that our method significantly improves both seizure detection performance and the quality of learned graph representations. 
\end{itemize}

The remainder of this article is organized as follows: Section~\ref{sec:sec2} reviews related works, including graph learning-based EEG seizure detection, and LLM applications in epilepsy diagnosis and treatment. Section~\ref{sec:sec3} introduces the proposed two-stage framework for graph structure learning and edge refinement. In Section~\ref{sec:sec4}, we conduct comprehensive experiments to evaluate seizure detection performance, and establish a benchmark for LLM-based clinical graph structure refinement. Finally, Section~\ref{sec:sec5} concludes this paper.

\section{Related Works} \label{sec:sec2}

\subsection{Graph Representation Learning for EEG Seizure Detection}

Graph representation learning has emerged as a powerful approach for EEG-based seizure detection, as it enables the modeling of complex, non-Euclidean spatial dependencies among EEG electrodes~\cite{bhandari2024exploring}. Early research primarily adopted static graph representations, where the connectivity between electrodes is predefined according to physical locations or simple statistical criteria~\cite{klepl2024graph}. For example,~\cite{ZHAO2021106277} constructs static graphs by calculating Pearson correlation matrices for EEG channel pairs and applying thresholds to define edges between highly correlated channels.~\cite{wang2020sequential} transforms EEG signals into the frequency domain using Fast Fourier Transform, treating each electrode as a node and establishing connections based on the Visibility Graph rule. Despite their simplicity, static graph-based approaches struggle to reflect the dynamic and evolving relationships among EEG channels, as they rely on fixed graph structures. This limitation has motivated the development of dynamic graph representation learning methods that more effectively capture the spatial-temporal nature of EEG networks.

Depending on how dynamic connectivity is defined and updated, existing dynamic graph learning strategies for EEG signals can be grouped into three categories: (i) Correlation-based Graph Construction, (ii) Learnable Graph Construction, and (iii) Generative Graph Construction. \textit{Correlation-based methods}\cite{tang2022selfsupervised,tao2022seizure,10380667} build dynamic graphs by computing channel correlations (e.g., Pearson or cross-correlation) over sliding time windows, but these only reflect surface-level statistical relationships and are easily affected by noise, resulting in unstable graph structures. \textit{Learnable graph construction methods}~\cite{huang2025eeg,li2024dynamical} enable adaptive graph connectivity by leveraging attention mechanisms or mask learning during training, which increases modeling flexibility. However, these methods typically lack explicit physiological grounding and treat graph evolution as a byproduct of optimization, making it difficult to accurately model the temporal dynamics of neural activity during seizures or to integrate domain knowledge about seizure propagation. \textit{Generative graph construction methods} embed graph generation modules within neural networks, dynamically synthesizing graph structures from latent EEG representations. Notable examples include self-supervised learning~\cite{ho2023self,weng2024self} and diffusion modeling approaches~\cite{chen2022brainnet,shu2024data,xu2026geogen}, which aim to model the spatial and temporal dynamics of EEG networks. While such data-driven methods are highly flexible and adaptive, they often lack physical interpretability and physiological plausibility, and may produce neural connections that are not biologically meaningful. Moreover, these approaches frequently overlook the inherent directionality and spatiotemporal aspects of neural signal propagation in epileptic seizures, thus limiting their ability to fully capture the underlying pathological mechanisms.

\subsection{LLM Applications in Epilepsy Diagnosis}

Existing LLM-based EEG seizure diagnosis methods can be broadly categorized into two types. The first category is \textbf{\textit{LLM-enhanced Clinical Text Analysis}}, which encompasses the majority of existing studies. These works require large amounts of natural language as training corpus, such as electronic medical records and expert diagnostic narratives. For instance, one early study leveraged LLMs to predict seizure recurrence risk in children by analyzing clinical notes~\cite{beaulieu2023predicting}; subsequent research introduced specialized models to interpret seizure semiology text for epileptogenic zone localization~\cite{yang2024episemollm}; other efforts explored automatic classification of seizure types from narrative case reports~\cite{kerr2025supervised}, as well as the extraction of seizure frequency and related attributes from clinical documentation using fine-tuned language models~\cite{abeysinghe2025leveraging}. More recent work assessed the performance of general-purpose LLMs like GPT series in distinguishing seizure types based on patient self-reported descriptions~\cite{ford2025can,mishra2025thought2text}. These approaches, however, demand not only EEG sequence data but also extensive auxiliary datasets and detailed expert annotations, which pose significant challenges for clinical implementation. Collecting such data and granular labels is costly and time-consuming, and the resulting large input size can considerably increase algorithmic inference time.

The second category encompasses a smaller set of research, which directly focuses on \textbf{\textit{EEG signal modeling with large-scale language models}}. In this line of work, researchers have proposed to use self-supervised pretraining~\cite{jiang2024large} to obtain EEG signal large models, developed neural tokenization methods to map EEG data into discrete embeddings compatible with LLMs~\cite{kim2024eeg}, and designed frameworks that integrate EEG feature extraction with prompt-based or token-based LLM inference~\cite{jiang2024neurolm,barmpas2025neurorvq}. While these approaches show promise, their main limitation is that the EEG data does not directly contribute to graph structure learning or optimizing the underlying graph representations. Instead, these methods primarily rely on tokenization and sequential modeling, missing opportunities to exploit the inherent spatial-temporal relationships within EEG signals.

\section{Methodology} \label{sec:sec3}

\subsection{Problem Definition and Overall Framework}
In this work, we propose a robust and interpretable approach for EEG-based seizure detection. The objective is to automatically determine whether a seizure event occurs within a signal window by capturing physiologically meaningful interactions among brain regions and filtering out noisy or redundant connections.

\begin{figure*}[!htbp]
\centering
\includegraphics[width=1.0\textwidth]{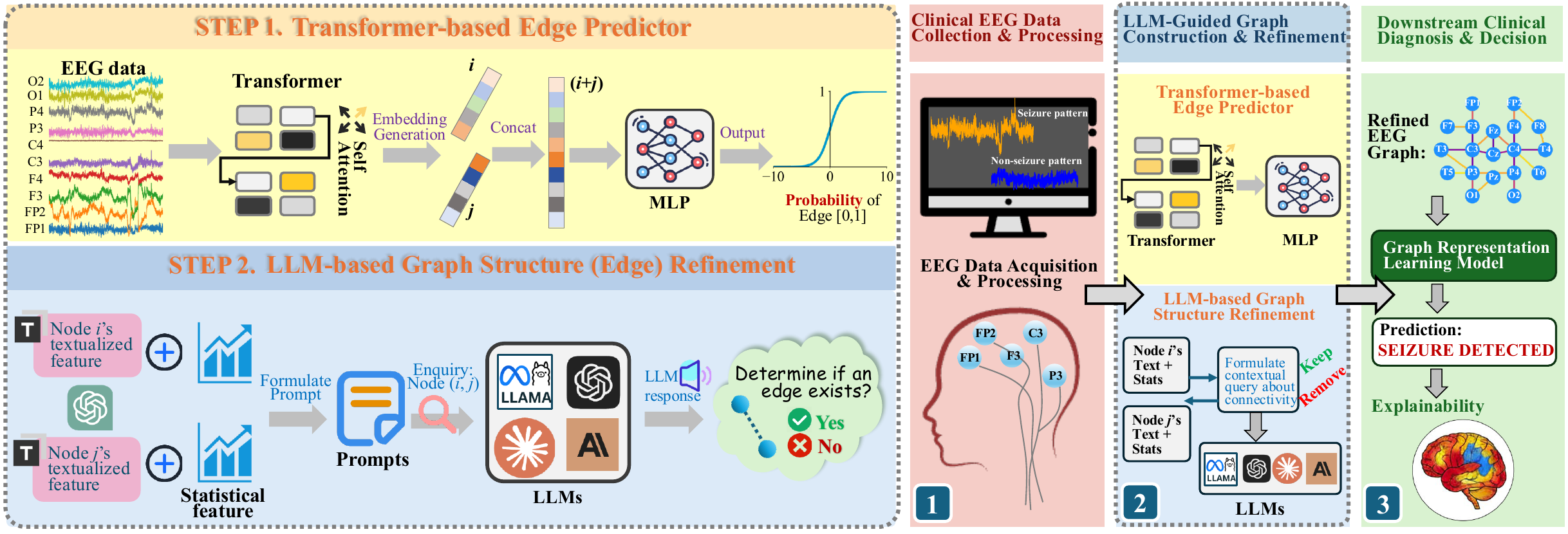}
\caption{(Left) The proposed two-stage LLM-based clinical graph generation and refinement framework. (Right) The overall clinical EEG seizure diagnosis pipeline, in which the proposed framework corresponds to Step 2 for constructing and refining the EEG graph before downstream seizure prediction and interpretation.}
\label{fig1_model_architecture}
\end{figure*}

Figure~\ref{fig1_model_architecture} (left panel) shows our proposed two-stage LLM-based clinical graph generation and refinement framework. The first stage employs a Transformer-based edge predictor to construct the preliminary graph structure from EEG data, modeling pairwise dependencies among channels. The second stage utilizes LLMs to perform context-aware edge refinement, ensuring that the resulting graph is sparse, consistent, and neurophysiologically plausible. The right panel illustrates the complete clinical workflow for EEG seizure diagnosis, in which the refined EEG graph is subsequently processed by a graph representation learning model for downstream seizure detection task, providing both enhanced accuracy and improved interpretability.

\subsection{Basic Graph Representation Learning using Transformer-based Edge Predictor}

In the first stage of our framework, we aim to construct an initial graph that captures the potential relationships between EEG channels. Given the sequential EEG data from $N$ channels, denoted as $\mathbf{X_{\tau}} = \{x_{\tau}^{1}, x_{\tau}^{2}, \ldots, x_{\tau}^{n}\}$, where $x_{\tau}^{i} \in \mathbb{R}^\tau$ represents the time series recorded at channel $i$ at time $\tau$, we employ a Transformer encoder to extract informative feature representations for each channel.

\noindent \textbf{Feature Encoding with Transformer:} Each EEG channel’s time series ${x}_{\tau}^{i}$ is first passed through a Transformer encoder, which models both local and global temporal dependencies within the signal. The output is an embedding vector for each node (channel), denoted as:
\begin{equation}
h_{\tau}^{i} = \text{Transformer}(x_{\tau}^{i}), \quad h_{\tau}^{i} \in \mathbb{R}^d
\label{eq:eq1}
\end{equation}

\noindent where $d$ is the dimension of the learned embedding.

\noindent \textbf{Edge Probability Prediction:} For each pair of channels $(i, j)$, we concatenate their embeddings $h_{\tau}^{i}$ and $h_{\tau}^j$ to form a joint representation $z_{\tau}^{ij}$:
\begin{equation}
z_{\tau}^{ij} = [h_{\tau}^i ; h_{\tau}^j] \in \mathbb{R}^{2d}
\label{eq:eq2}
\end{equation}

This concatenated feature is then fed into a two-layer MLP to predict the probability that an edge connection exists between channel $i$ and channel $j$:
\begin{equation}
p_{\tau}^{ij} = \sigma\left( \mathbf{w}_2^\top \, \phi\left( \mathbf{w}_1 z_{\tau}^{ij} + \mathbf{b}_1 \right) + b_2 \right) \tag{3}
\label{eq:eq3}
\end{equation}

\noindent where $\sigma(\cdot)$ is the sigmoid activation function, $\phi(\cdot)$ is the ReLU activation, $\mathbf{w}_1$ and $\mathbf{w}_2$ are the weight matrices for the 1st and 2nd layers, $\mathbf{b}1$ and $b_2$ are the corresponding biases, and $p_{\tau}^{ij} \in [0, 1]$ is the predicted edge existence probability between nodes $i$ and $j$.

\noindent \textbf{Initial Edge Set Selection:} We apply a threshold $\varphi$ to determine the presence of an edge:
\begin{equation}
e_{\tau}^{ij} = \begin{cases} p_{\tau}^{ij}, & \text{if } p_{\tau}^{ij} > \varphi \\ 0, & \text{otherwise} \end{cases}
\end{equation}

\noindent The resulting binary adjacency matrix $E = [e_{ij}]$ defines the initial edge set for the constructed graph.

In summary, the first stage generates a data-driven initial graph structure by leveraging the temporal modeling capabilities of Transformer and the discriminative power of MLP for edge prediction. The initial graph will be further optimized in the next stage using an LLM-based refinement mechanism.

\subsection{LLM-based Graph Structure Refinement}
The second stage of the proposed framework aims to further improve the quality and interpretability of EEG graph structure by refining the edge set generated in the first step using large language models. The motivation for this approach lies in the ability of LLMs to reason over both statistical and semantic information, allowing them to make informed decisions about whether an edge between two EEG channels is meaningful or redundant. This process addresses the limitations of existing methods, which typically lack explicit mechanisms for context-aware edge selection and noise reduction.

For each candidate edge $(i,j)$ in the initial graph constructed in the first step, we design proprietary prompts to query the LLM. These prompts incorporate both the textualized and statistical features of nodes $i$ and $j$, ensuring that the model considers both context and data-driven evidence. The proprietary prompt consists of two parts: (a) the textualized description of the EEG data at the current time window using natural language. (b) summarized statistical features of the studied EEG data at the current time window.

\noindent \textbf{Textualized EEG Feature Construction:} For each node, we employ GPT-4o to describe its relevant characteristics of the EEG signal as natural language. For node $i$, the textualized feature includes: (a) The channel label (e.g., ``F3''). (b) The anatomical location (e.g., ``frontal lobe''). (c) High-level observations from the EEG data sequence (e.g., ``shows sharp spikes'', "background is normal"), and the changing trend of the EEG data sequence. A sample textual description for node $i$ is given here:

\begin{tcolorbox}[colback=gray!5!white, colframe=lightblueframe, left=1pt, right=1pt, top=1pt, bottom=1pt, title=Example of Textualized EEG Feature Construction]
Channel F3, located in the frontal lobe, shows moderate amplitude with intermittent sharp spikes. The EEG signal exhibits an increasing trend in amplitude over the observed time window.
\end{tcolorbox}

\noindent \textbf{EEG Statistical Feature Construction:} Following the common practice in EEG signal statistics that are crucial and most discriminative for seizure detection~\cite{Zhang2022A,Shankar2021Classification,Sharmila2020Evaluation}, we extract the the statistical features listed below using Python:
(a) \textit{Mean amplitude:} The average value of the EEG signal over the selected time window, reflecting the baseline activity level. (b) \textit{Standard deviation:} The variability or dispersion of the EEG signal amplitudes, indicating the spreading of the signals. (c) \textit{Dominant frequency:} The most prominent frequency component in the EEG signal, showing the primary oscillatory activity. (d) \textit{Minimum:} The lowest amplitude observed in the EEG segment, capturing extreme negative fluctuations. (e) \textit{Maximum:} The highest amplitude observed in the EEG segment, capturing extreme positive fluctuations. (f) \textit{Median:} The middle value of the sorted EEG amplitudes, representing the central tendency. (g) \textit{(q25):} The value below which 25\% of the EEG data points fall, reflecting the lower distribution. (h) \textit{(q75):} The value below which 75\% of the EEG data points fall, indicating the upper distribution. (i) \textit{Skewness:} A measure of the asymmetry in the amplitude distribution of the EEG signal. (j) \textit{Kurtosis:} A measure of the peakedness or flatness in the amplitude distribution of the EEG signal. (k) \textit{Energy:} The total power of the EEG signal, calculated as the sum of squared amplitudes. (l) \textit{Zero-crossing rate:} The number of times the EEG signal crosses the zero axis, often related to its frequency content.

These statistics are computed over the relevant EEG time window for each channel and presented in a standardized textual format. For example:
\begin{tcolorbox}[colback=gray!5!white, colframe=lightblueframe, left=1pt, right=1pt, top=1pt, bottom=1pt, title=Example of EEG Statistical Feature Construction]
Mean amplitude: 12.3 $\mu V$, Std: 4.5 $\mu V$, Dominant frequency: 8.5 Hz, Min: -6.2 $\mu V$, Max: 22.7 $\mu V$, Median: 10.1 $\mu V$, Q25: 7.8 $\mu V$, Q75: 14.5 $\mu V$, Skewness: 0.32, Kurtosis: 2.1, Energy: 1532.6, Zero-crossing rate: 42.
\end{tcolorbox}

\noindent \textbf{Prompt Design for Edge Refinement:} The full prompt for a candidate edge $(i,j)$ is constructed by concatenating the textual and statistical features of both nodes and then forming a specific inquiry. Suppose EEG channel F3 and T4 to be node $i$ and $j$, respectively. Then the LLM-based edge refinement prompt is given in the following:
\begin{tcolorbox}[breakable, enhanced, colback=gray!5!white, colframe=lightorange, left=1pt, right=1pt, top=1pt, bottom=1pt, title=LLM-generated contextual prompt]
Node $i$: Channel F3, located in the frontal lobe, shows moderate amplitude with intermittent sharp spikes. The EEG signal exhibits an increasing trend in amplitude over the observed time window. Mean amplitude: 12.3 $\mu V$, Std: 4.5 $\mu V$, Dominant frequency: 8.5 Hz, Min: -6.2 $\mu V$, Max: 22.7 $\mu V$, Median: 10.1 $\mu V$, Q25: 7.8 $\mu V$, Q75: 14.5 $\mu V$, Skewness: 0.32, Kurtosis: 2.1, Energy: 1532.6, Zero-crossing rate: 42.

Node $j$: Channel T4, located in the temporal lobe, shows high amplitude with pronounced rhythmic slowing. The EEG signal displays a stable pattern with occasional bursts throughout the observed time window. Mean amplitude: 18.2 $\mu V$, Std: 5.1 $\mu V$, Dominant frequency: 4.2 Hz, Min: -3.1 $\mu V$, Max: 25.8 $\mu V$, Median: 17.4 $\mu V$, Q25: 15.0 $\mu V$, Q75: 20.2 $\mu V$, Skewness: -0.15, Kurtosis: 2.7, Energy: 2018.3, Zero-crossing rate: 33.

Question: Does a meaningful functional connection exist between Node $i$ and Node $j$ in this EEG segment? Answer yes or no, based on the context above.
\end{tcolorbox}

This prompt is submitted to the LLM (e.g., GPT-5-mini), which returns a binary response ("yes" or "no") to indicate whether the edge should be retained. By iterating this process over all candidate edges, we obtain the final refined adjacency matrix where only edges validated by the LLM are preserved. This LLM-based refinement process enables more intelligent and interpretable graph construction, filtering out spurious and redundant edges based on a combination of contextual reasoning and data-driven evidence. Algorithm~\ref{alg:two_stage_framework} illustrates the overall procedure for LLM-based edge refinement.

\section{Experiments} \label{sec:sec4}
In this section, we aim to answer the following research questions: \textit{\textbf{RQ1:}} {How do different graph construction and refinement strategies influence downstream EEG seizure detection performance?} \textit{\textbf{RQ2:}} {How do various general-purpose LLMs perform as structural judges in graph reasoning?} \textit{\textbf{RQ3:}} {What are the interpretability and consistency characteristics of LLM-based graph reasoning in clinical EEG contexts?

\subsection{Dataset and Experimental Setup}
\noindent \textbf{Dataset.} We conduct experiments on the publicly available EEG seizure diagnosis benchmark: Temple University Hospital EEG Seizure Corpus (TUSZ) v1.5.2~\cite{shah2018temple}. TUSZ is currently one of the largest clinical EEG corpora for seizure diagnosis, comprising 5,612 EEG recordings and 3,050 seizure annotations. It includes 19 EEG channels recorded using the standard 10-20 system, covering a broad range of subjects across diverse clinical conditions. For interpretability analysis, we leverage the seizure onset-offset labels and event type annotations provided with TUSZ dataset.

\noindent \textbf{Baseline Methods.} To ensure a fair and comprehensive comparison, all models are built upon the same \textbf{modified GraphS4mer} backbone~\cite{tang2023modeling}, where the original internal graph structure learning module is disabled and replaced by externally supplied graph structures. The detailed backbone configuration is in Appendix~\ref{app:graphs4mer_backbone}. We consider two categories of baselines: 
\textbf{(1) \textit{Conventional graph construction methods}}, including (a) \textit{Self-Correlation Graph}, which constructs graphs based on pairwise channel correlations; (b) \textit{Distance-based Graph}, which connects channels according to their physical distance; (c) \textit{Attention-based Graph}, which utilizes attention mechanisms to learn adaptive edge weights; (d) \textit{K-Nearest Neighbor (KNN) Graph}, where each node is connected to its $k$ most similar channels based on signal similarity; (e) \textit{Generative Module-based Graph}, which learns graph structures through generative neural network modules. \textbf{(2) \textit{LLM-based graph structure refinement methods}}, where we benchmark a range of state-of-the-art general-purpose LLMs as structural judges, including Mistral 7B, Mistral 8×7B, Llama 3.1-70B, Llama 3.2-90B, Gemini-2.5-Pro, Gemini-2.5-Flash, GPT-4o, GPT-4.1, GPT-5, GPT5-mini. Each model independently refines the graph produced by the Transformer-based edge predictor. 


\begin{algorithm}[ht]
\footnotesize
\caption{Two-stage LLM-based Graph Structure Learning for EEG Seizure Detection}
\label{alg:two_stage_framework}
\begin{algorithmic}[1]
\Require Multichannel EEG data $\mathbf{X}_{\tau} = \{x_{\tau}^{1}, x_{\tau}^{2}, \ldots, x_{\tau}^{n}\}$; threshold $\varphi$ for edge selection; large language model $\mathcal{M}_{LLM}$;
\Ensure Final refined adjacency matrix $\mathbf{E}_{\tau}^*$;
\State \textbf{Step 1: Initial Graph Construction}
\For{each channel $i$}
    \State Obtain node embedding $h_{\tau}^{i} = \mathrm{Transformer}(x_{\tau}^{i})$
\EndFor

\For{each channel pair $(i,j)$}
    \State Concatenate embeddings: $z_{\tau}^{ij} = [h_{\tau}^i; h_{\tau}^j]$
    \State Compute edge probability: $p_{\tau}^{ij} = \sigma(\mathrm{MLP}(z_{\tau}^{ij}))$
    \If{$p_{\tau}^{ij} > \varphi$}
        \State Set $e_{\tau}^{ij} = p_{\tau}^{ij}$ \ {(edge weight in initial graph)}
    \Else
        \State Set $e_{\tau}^{ij} = 0$ \ {(no edge connection)}
    \EndIf
\EndFor
\State Obtain initial adjacency matrix $\mathbf{E}_{\tau}^{ini}$ from all $e_{\tau}^{ij}$

\State \textbf{Step 2: LLM-based Edge Refinement}
\State Initialize $\mathbf{E}_{\tau}^* \gets \mathbf{E}_{\tau}^{ini}$
\For{each node pair $(i,j)$ with $e_{\tau}^{ij} > 0$ in $\mathbf{E}_{\tau}^{ini}$}
    \State Construct textualized and statistical features for node $i$ and node $j$
    \State Formulate prompt $\mathcal{Q}_{\tau}^{ij}$ from features of node $i$ and node $j$
    \State Submit $\mathcal{Q}_{\tau}^{ij}$ to $\mathcal{M}_{LLM}$ and obtain binary response $R_{\tau}^{ij}$
    \If{$R_{\tau}^{ij}$ is ``yes''}
        \State Set $e_{\tau}^{ij} = p_{\tau}^{ij}$
    \ElsIf{$R_{\tau}^{ij}$ is ``no''}
        \State Set $e_{\tau}^{ij} = 0$
    \EndIf
\EndFor
\State \Return $\mathbf{E}_{\tau}^*$
\end{algorithmic}
\end{algorithm}

\subsection{Results and Performance Comparison}

\begin{table}[!htbp]
\centering
\footnotesize
\caption{Seizure detection performance under different graph construction / refinement methods.}
\rowcolors{2}{gray!20}{white}
\resizebox{\linewidth}{!}{
\begin{threeparttable}
\begin{tabular}{lccc}
\toprule
\textbf{Model} & \textbf{F1} & \textbf{Accuracy} & \textbf{Recall} \\
\midrule
\multicolumn{4}{l}{\textit{None LLM-based Graph Construction Methods}} \\
Distance-based $\mathcal{G}$                   & 0.6508 & 0.8225 & 0.7049 \\
Self-Correlation $\mathcal{G}$         & 0.6682 & 0.8453 & 0.7158 \\
KNN-based $\mathcal{G}$                        & 0.7156 & 0.8762 & 0.7362 \\
Attention-based $\mathcal{G}$                  & 0.7338 & 0.8894 & 0.7537 \\
\textbf{Generative Graph Learner }$\boldsymbol{\mathcal{G}}$  & \underline{0.7351} & \underline{0.9039} & \underline{0.7625} \\
\midrule
\multicolumn{4}{l}{\textit{LLM-based Graph Structure Refinement Methods}} \\
Mistral 7B + Transformer $\mathcal{G}$         & 0.7458 & 0.9087 & 0.7705 \\
Mistral 8×7B + Transformer $\mathcal{G}$       & 0.7493 & 0.9056 & 0.7784 \\
Llama 3.1-70B + Transformer $\mathcal{G}$      & 0.7522 & 0.9196 & 0.7786 \\
Llama 3.2-90B + Transformer $\mathcal{G}$      & 0.7587 & 0.9173 & 0.7815 \\
Gemini 2.5-Flash + Transformer $\mathcal{G}$   & 0.7613 & 0.9207 & 0.7836 \\
Gemini 2.5-Pro + Transformer $\mathcal{G}$     & 0.7685 & 0.9261 & 0.7879 \\
GPT-4o + Transformer $\mathcal{G}$             & 0.7726 & 0.9278 & 0.7947 \\
GPT-4.1 + Transformer $\mathcal{G}$            & 0.7792 & 0.9236 & 0.7933 \\
GPT-5-mini + Transformer $\mathcal{G}$          & 0.7859 & 0.9275 & 0.8024 \\
\textbf{GPT-5 + Transformer} $\boldsymbol{\mathcal{G}}$ & \textbf{0.7907} & \textbf{0.9315} & \textbf{0.8058} \\
\bottomrule
\end{tabular}
\begin{tablenotes}
\item ${*}${\scriptsize All models share the same GraphS4mer backbone.}
\end{tablenotes}
\end{threeparttable}
}
\label{tab:task_level_eval}
\end{table}

\noindent \textbf{Experimental Setup.} To evaluate the influence of different graph construction and refinement strategies on downstream seizure detection, we employ GraphS4mer as a unified backbone across all experiments, ensuring that performance variations arise solely from graph structural differences. The upstream graph inputs include two categories: (1) Non-LLM-based graph construction methods and (2) LLM-based graph structure refinement methods, where various LLMs serve as structural judges to refine the initial constructed graphs by Transformer model. The performance is evaluated using three widely adopted metrics: \textit{\textbf{F1-score}}, \textit{\textbf{Accuracy}}, and \textit{\textbf{Recall}}.

\noindent\textbf{Results and Analysis.} Table~\ref{tab:task_level_eval} summarizes the seizure detection results across all baselines. Overall, models employing LLM-based graph refinement consistently outperform all non-LLM graph construction methods, demonstrating the effectiveness of leveraging language-driven reasoning to refine EEG connectivity structures. Among traditional approaches, Generative Graph Learner achieves the highest F1, indicating that learning-based graph formation offers moderate benefits over handcrafted correlations. In contrast, integrating different LLMs yields significant performance improvements, with GPT-5 + Transformer $\mathcal{G}$ achieving the best results (F1 = 0.7907, Accuracy = 0.9315, Recall = 0.8058). This performance gain highlights that larger, instruction-tuned models possess stronger reasoning and consistency when determining clinically meaningful connections between EEG channels. The steady improvement from smaller models (e.g., Mistral 7B) to larger ones (e.g., Llama 3.2-90B, GPT-5) further suggests that graph reasoning capability scales with model capacity, validating the efficacy of our proposed LLM-based graph refinement paradigm.



\subsection{LLM Clinical Graph Reasoning Benchmark}

\noindent \textbf{Experimental Setup.} To systematically evaluate the reasoning and structural refinement capabilities of different LLMs, we establish a unified benchmark named LLM Clinical Graph Refinement Benchmark. In this setting, each LLM independently serves as a structural judge that refines the graph generated by Transformer-based edge predictor under identical conditions, ensuring fairness and reproducibility. We benchmark a diverse set of representative models across different scales and architectures, including Mistral 7B, Mistral 8×7B, Llama 3.1-70B, Llama 3.2-90B, Gemini 2.5-Flash, Gemini 2.5-Pro, GPT-4o, GPT-4.1, GPT-5, and GPT-5-mini. Beyond the downstream seizure detection metrics (F1, Accuracy, and Recall), we further compute graph-level statistics: (I) \textbf{\textit{Graph Sparsity}}~\cite{cini2023sparse} and (II) \textbf{\textit{Jensen-Shannon Divergence(JSD)}}~\cite{wang2025jensen} to quantify each model's reasoning consistency and structural diversity. 

\begin{table}[!htbp]
\centering
\footnotesize
\caption{Graph-level reasoning characteristics of different LLMs.}
\rowcolors{2}{gray!20}{white}
\resizebox{0.94\linewidth}{!}{
\begin{tabular}{lccc}
\toprule
\textbf{Model} & \textbf{Graph Sparsity} & \textbf{JSD} \\
\midrule
Mistral 7B + Transformer $\mathcal{G}$         & 0.4213 & 0.0576\\
Mistral 8×7B + Transformer $\mathcal{G}$       & 0.4167 & 0.0532 \\
Llama 3.1-70B + Transformer $\mathcal{G}$      & 0.4019 & 0.0442 \\
Llama 3.2-90B + Transformer $\mathcal{G}$      & 0.3984 & 0.0407 \\
Gemini 2.5-Flash + Transformer $\mathcal{G}$   & 0.3931 & 0.0369 \\
Gemini 2.5-Pro + Transformer $\mathcal{G}$     & 0.3875 & 0.0335 \\
GPT-4o + Transformer $\mathcal{G}$             & 0.3826 & 0.0314 \\
GPT-4.1 + Transformer $\mathcal{G}$            & 0.3778 & 0.0306 \\
GPT-5-mini + Transformer $\mathcal{G}$          & 0.3758 & 0.0289 \\
\textbf{GPT-5 + Transformer} $\boldsymbol{\mathcal{G}}$    & \textbf{0.3741} & \textbf{0.0265}\\
\bottomrule
\end{tabular}
}
\label{tab:LLM_graph_bench}
\end{table}

\noindent \textbf{Results and Analysis.} Table~\ref{tab:LLM_graph_bench} reports the benchmark results across all models. Overall, the proposed benchmark reveals distinct reasoning behaviors and scaling trends among the evaluated LLMs. Smaller models such as Mistral 7B and Mistral 8×7B achieve moderate seizure detection accuracy but exhibit relatively high graph diversity, indicating less stable structural reasoning.
In contrast, mid-sized open-source models like Llama 3.1-70B and Llama 3.2-90B improve both F1 and structural consistency, reflecting more coherent edge judgment. Gemini 2.5-Pro and GPT-series models demonstrate the strongest reasoning reliability and overall detection performance, with GPT-5 achieving the highest F1 score, edge agreement rate, and lowest divergence. These results suggest that reasoning capability in graph refinement scales with model capacity and training diversity. Moreover, we observe a clear trade-off: smaller models tend to explore more novel edge connections (higher diversity but lower stability), whereas larger models yield more clinically plausible and consistent graph structures. This benchmark thus provides a standardized lens through which to compare LLMs' reasoning quality in clinical graph-based tasks.

\subsection{Interpretability \& Consistency Analysis}
To further understand how different LLMs perform structural reasoning during EEG graph refinement, we conduct interpretability and consistency analysis.
We visualize node-level importance distributions and assess the physiological plausibility of the refined graphs with respect to clinical EEG knowledge and ground-truth annotations.
\begin{figure}[!htbp]
\centering
\includegraphics[width=1.0\columnwidth]{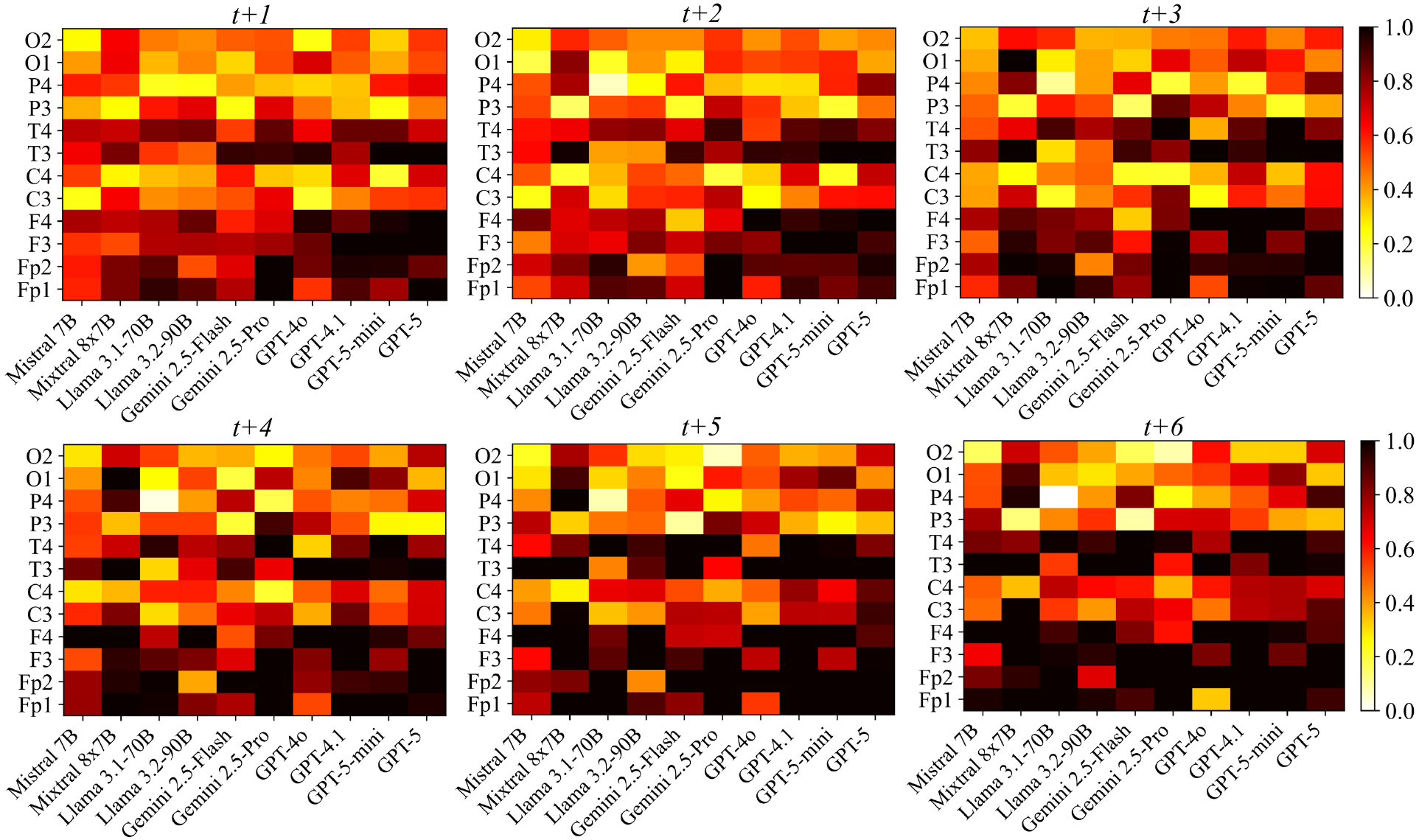}
\caption{Node feature importance distributions across different LLM-based graph refiners in continuous time steps, illustrating seizure onset in frontal region and propagation toward temporal area.}
\label{fig_RQ3_node_feature_importance}
\end{figure}

Figure~\ref{fig_RQ3_node_feature_importance} illustrates the temporal evolution of node-level importance distributions across different LLM-based graph refiners during a seizure episode. As shown from t+1 to t+6, the activation pattern exhibits a physiologically consistent propagation from frontal (Fp1, Fp2, F3, F4) to temporal (T3, T4) regions, reflecting the typical onset and spread of epileptic activity. Among all models, the GPT family demonstrates more stable and spatially focused activations over key cortical areas, indicating stronger interpretability and better alignment with clinically meaningful EEG dynamics. In contrast, models such as Mistral and Llama display scattered and less consistent node activations. Notably, GPT-5 achieves the most coherent and biologically plausible feature patterns, followed by GPT-5-mini and GPT-4.1, corroborating the superior reasoning and refinement capability of advanced LLMs in capturing seizure-related brain connectivity.

\begin{figure}[!htbp]
\includegraphics[width=0.85\columnwidth]{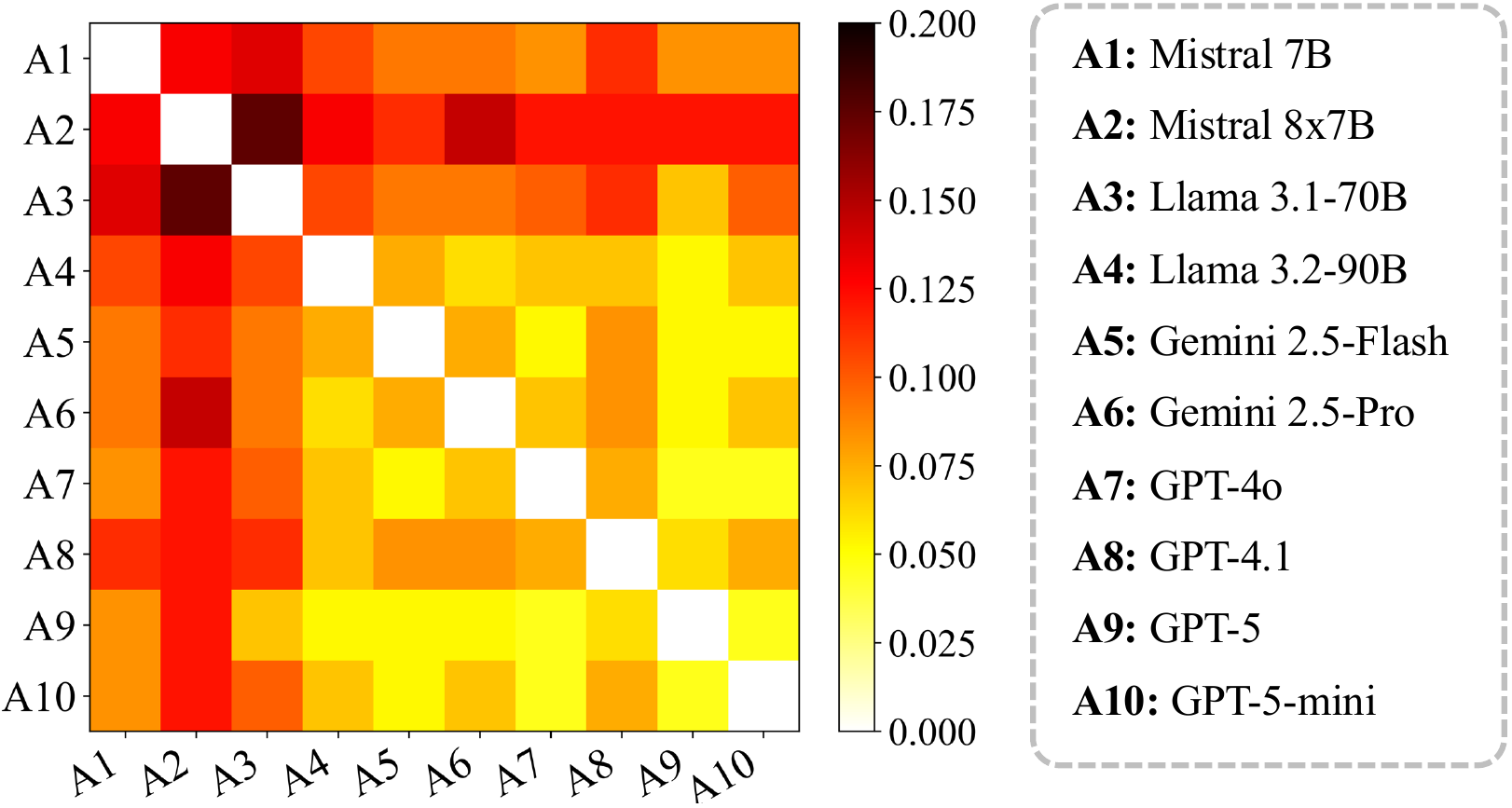}
\caption{Edge Difference Heatmap. It shows the pairwise structural discrepancies among the refined graphs from different LLMs.}
\label{fig_RQ3_edge_difference}
\end{figure}

Figure~\ref{fig_RQ3_edge_difference} shows the pairwise edge difference heatmap of LLM-based graph refinement. Brighter regions (bottom-right) indicate higher structural consistency, while darker tones (upper-left) reveal greater divergence across models. Specifically, GPT model series (GPT-4o, GPT-4.1, GPT-5, GPT-5-mini) exhibit the highest internal consistency, as indicated by the lighter color in the lower-right corner, suggesting that these models generate highly similar and stable graph structures. This consistency implies that GPT models share a coherent reasoning pattern when inferring physiologically meaningful brain connectivity. In contrast, the Mistral and Llama series show substantially higher pairwise differences, reflecting less stable and more stochastic refinement behaviors. The Gemini models lie in between, demonstrating moderate consistency with the GPT family. Overall, the gradual reduction in structural discrepancy from Mistral to GPT-5 aligns with the progression of reasoning capability, indicating that more advanced LLMs achieve not only higher accuracy in edge refinement but also improved physiological plausibility in representing seizure-related brain networks.

\begin{figure}[!htbp]
\includegraphics[width=1.0\columnwidth]{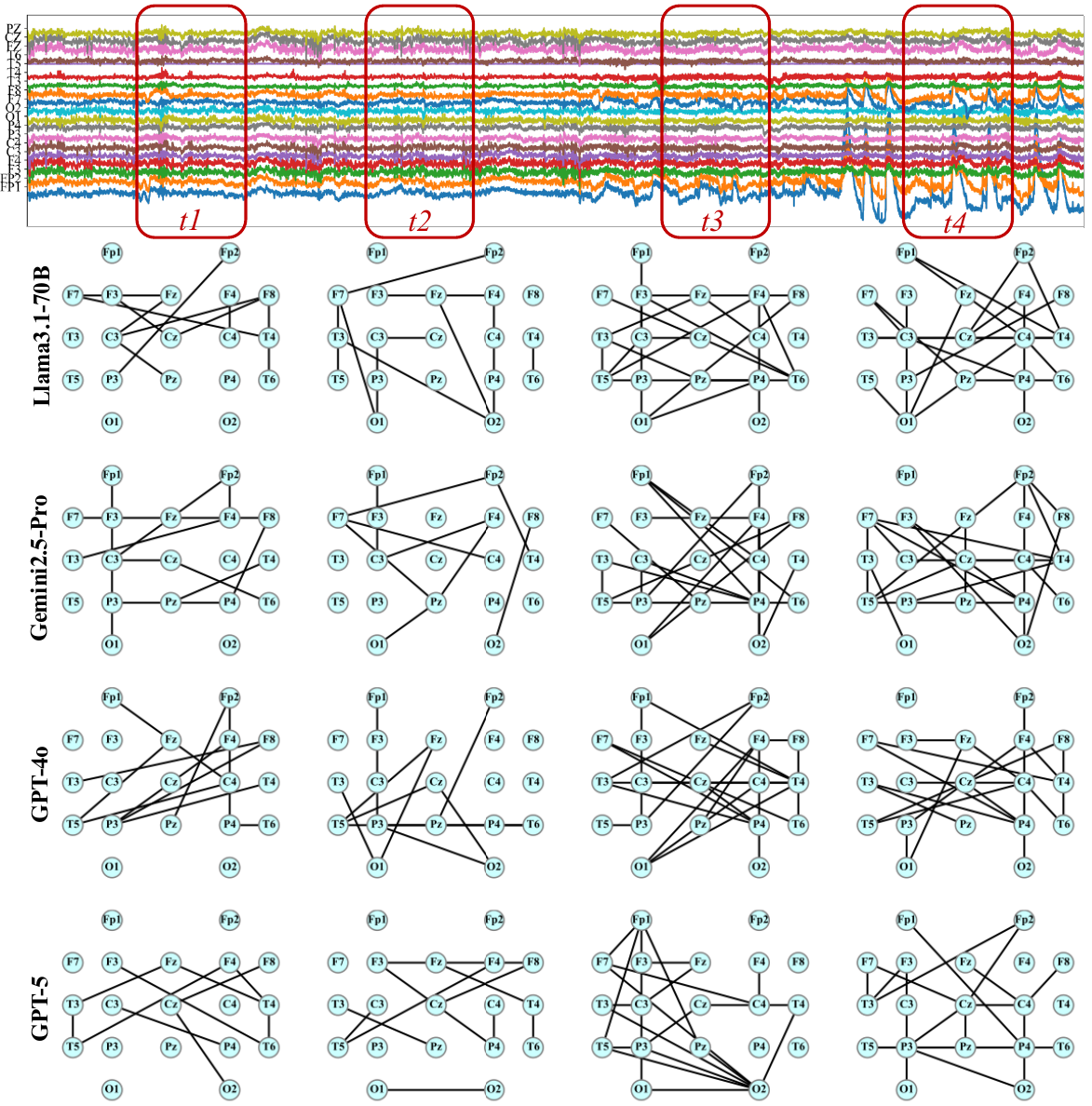}
\caption{Visualization of the spatiotemporal evolution of LLM-guided EEG graphs.}
\label{fig_RQ3_case_study}
\end{figure}

Figure~\ref{fig_RQ3_case_study} presents a case study comparing the graphs refined by different LLMs across four time windows of a standard EEG signal sequence. The EEG signals exhibit a clear propagation pattern from frontal to temporal, central, and finally parietal–occipital regions, which reflects the typical spatial evolution of seizure activity. Consistent with this physiological trend, GPT series (GPT-4o and GPT-5) produce structured and spatially coherent graphs that capture this progressive propagation while maintaining sparse and physiologically plausible edge connectivity. Gemini 2.5-Pro demonstrates moderate consistency but introduces redundant inter-regional links, suggesting moderate precision in edge reasoning. Although graphs refined by Llama 3.1-70B reflect the seizure propagation pattern, it yields more dense and noisy connections. Overall, these results highlight that higher-capacity LLMs not only infer more accurate and stable graph topologies but also better align with the underlying neurophysiological dynamics observed in real EEG data.

\section{Conclusions} \label{sec:sec5}
In this paper, we propose a novel two-stage framework for EEG seizure detection that leverages the reasoning and contextual understanding capabilities of Large Language Models (LLMs) for clinical graph structure refinement. By integrating a Transformer-based edge predictor with an LLM-driven edge refinement module, our approach effectively identifies and removes redundant or irrelevant graph connections, improving graph quality and interpretability. We further establish an LLM-based clinical graph structure reasoning benchmark evaluating multiple general-purpose LLMs as graph structure judges. Extensive experiments on TUSZ benchmark demonstrate that our framework achieves superior seizure detection performance and generates more meaningful graph representations compared to existing methods. These findings highlight the potential of LLMs to bring new intelligence and interpretability to graph learning in neurophysiological signal analysis. Future work will explore domain-adaptive prompting and multimodal integration to further enhance interpretability, robustness, and generalization across broader biomedical graph learning tasks.


\bibliographystyle{named}
\bibliography{ijcai26}

\clearpage
\appendix

\section{GraphS4mer Backbone Configuration}
\label{app:graphs4mer_backbone}

GraphS4mer~\cite{tang2023modeling} is originally designed as a spatiotemporal graph neural network for multivariate biosignal classification. Its architecture consists of three main components: (1) stacked S4 layers for modeling long-range temporal dependencies within each sensor channel, (2) an internal graph structure learning (GSL) module for dynamically estimating adjacency matrices from learned temporal embeddings, and (3) GNN layers followed by temporal and graph pooling modules for downstream classification.

In our experimental evaluations, GraphS4mer is employed as the unified downstream representation learning backbone. To isolate the effect of different graph construction and refinement strategies, we modify the original GraphS4mer by disabling its internal GSL module. Specifically, the self-attention-based adjacency estimation, KNN graph mixing, and threshold pruning in the original GraphS4mer are removed. Instead, the adjacency matrix used by the GNN layers is externally provided by the graph construction/refinement method in our experiments.

Under this setting, the retained components of GraphS4mer include the S4 temporal encoder, the GNN spatial encoder, the temporal pooling layer, the graph pooling layer, and the final classifier. Given an EEG segment, the S4 encoder first extracts long-range temporal representations for each EEG channel. The externally supplied graph then defines the channel-level connectivity used by the GNN layers to perform spatial message passing. The resulting node and temporal representations are aggregated through pooling layers and passed to the classifier for seizure detection. This configuration ensures that all compared methods share the same downstream model capacity and training pipeline. The only varying factor is the input graph structure, including conventional graph construction methods and LLM-based graph refinement methods. Therefore, the only difference in downstream seizure detection performance is directly attributed to the quality of the constructed or refined EEG graph rather than to architectural differences in the backbone model.

\end{document}